\relax
\documentclass[letterpaper]{article} 
\usepackage{times}  
\usepackage{helvet} 
\usepackage{courier}  
\usepackage[hyphens]{url}  
\usepackage{graphicx} 
\usepackage{graphicx}  
\usepackage{algorithm}
\usepackage{algorithmicx}
\usepackage{algpseudocode}
\usepackage{appendix}
\usepackage{cite}
\usepackage{caption}
\usepackage{subfigure}
\usepackage{hyperref}

\title{Adjust Planning Strategies to Accommodate Reinforcement Learning Agents}
\author{\Large \textbf{Xuerun Chen}\\ 
National University of Defense Technology\\
chenxuerun18@nudt.edu.cn 
}
\begin{document}

\maketitle

\begin{abstract}
In agent control issues, the idea of combining reinforcement learning and planning has attracted much attention. Two methods focus on micro and macro action respectively. Their advantages would show together if there is a good cooperation between them. An essential for the cooperation is to find an appropriate boundary, assigning different functions to each method. Such boundary could be represented by parameters in a planning algorithm. In this paper, we create an optimization strategy for planning parameters, through analysis to the connection of reaction and planning; we also create a non-gradient method for accelerating the optimization. The whole algorithm can find a satisfactory setting of planning parameters, making full use of reaction capability of specific agents.\footnote{The code is available on https://github.com/chenxuerun/APS}
\end{abstract}

\section{Introduction}
The solution of many continuous decision problem can be described as such a process: agent set out from the initial state, then go through a series of intermediate state and finally reach the goal state. Imagine an agent in a maze, which needs to find some key positions and pass through them one by one to get out.

Agent has two types of behavior: one is the micro action taken at every state, which is similar to muscle activity, called \emph{reaction}; another is the change of trend in reactions taken over a period of time, which is similar to thought of human, called \emph{planning} \cite{SUTTON1999181}. For the agent in maze, reaction can be its every little moving step and planning can be its every determination of the position it should reach next.

In a \emph{complicated} scene with high-dimensional data stream, long-term decision process and sparse supervision signal, an agent trained only to react \cite{dqn,ddpg} can hardly perform well (See Appendix A for demonstration). However, combining reaction and planning \cite{prmrl,sptm,sorb} can significantly improve its capability.

The essence of such improvement is that agent has limited reaction capability and the introduction of planning releases agent from reacting in the whole task. If the agent in maze only know how to reach a nearby position, with consecutive adjacent positions given by a planner, it still has the ability to reach a specified remote position.

What is the connection between reaction and planning? Improving reaction capability means spending much time training a connecting structure \cite{bp,nn} before task. Planning is not required once the reaction capability reach a certain level, which is difficult in a complicated scene (See Appendix B for demonstration). Improving planning capability means designing a planner that can provide useful information for reacter or divide original task into easier tasks. Planning would consume much more resources during the task.

Considering the features of reaction and planning, there should be a way to make full of their advantages: giving agent enough reaction capability without consuming much resources before the task, and, ensuring well performance without consuming much resources during the task. To achieve this, an evaluation of the reaction capability is necessary, helping to get a compatible planner.

A recent work (SoRB)\cite{sorb} showed a novel way to handle problems in a complicated scene: agent first samples states as \emph{waypoints}, next connects waypoints to get a \emph{planning graph}, then finds a shortest path in the graph, and finally reacts along waypoints on the shortest path. They found a powerful tool to incorporate planning techniques into RL: distance estimates obtained from RL.

Based on SoRB, this paper analyze the effect of two \emph{planning parameters} on planning: the number of waypoints in and the maximum edge length of the planning graph. An \emph{online adapting algorithm} is then proposed, which can adjust the planning parameters base on complexity of state place and reaction capability of agent. With this algorithm, task will be handled with relatively little computational cost and high success rate.

\section{Background}

Before using online adapting algorithm, agent has obtained a reactive policy trained by RL and got a planner which constructs a planning graph and uses Dijkstra's Algorithm to find shortest path, dividing original task by setting subgoals along the path. Two parameters as mentioned above have significant effect on performance. An optimization strategy is created to find a satisfactory setting of the parameters. An modified pattern search method is created to accelerate optimization.\\

\noindent\emph{Goal-Conditioned RL}: The state of agent is determined by its current and goal state: $s,g\in \mathcal{S}$. At every state agent takes a reaction: $a\in \mathcal{A}$. It has a reactive policy: $\pi\in\mathcal{S}\times\mathcal{S}\rightarrow\mathcal{A}$. The environment of agent has a reward function: $r\in \mathcal{S}\times\mathcal{S}\times\mathcal{A}\rightarrow R$, and a transition function: $P\in \mathcal{S}\times\mathcal{S}\times\mathcal{A}\rightarrow\mathcal{S}\times\mathcal{S}$. The reactive policy is learned by DDPG \cite{ddpg} algorithm. Agent has a function that assess values of each pair of state and action: $Q\in \mathcal{S}\times\mathcal{S}\times\mathcal{A}\rightarrow R$. \emph{Ideally}, $Q(s_0,g,a_0)=E^\pi[\sum_{t=0}^{\infty}\gamma^tr(s_t,g,a_t)]$, where $\gamma\in(0,1)$ is a discount factor. By decreasing Bellman Error: $|Q(s,g,a)-r(s,g,a)-\gamma \max \limits_{a'} Q(s',g,a')|$, $Q$ values will approach to the ideal ones. By choosing reaction of larger $Q$ value: $\pi(s,g)=\max\limits_{a}Q(s,g,a)$, better performance can be achieved. After alternately optimizing $Q$ and $\pi$, agent will get better reacting and evaluating capability.\\

\noindent\emph{Distance Estimates Obtained from RL}: In order to construct a planning graph, agent must estimate distances between every pair of waypoints without additional information. If the environment has a special setting: $r(s,g,a)=-1$ and $\gamma=1$ \cite{Kaelbling1993LearningTA}, then DDPG algorithm will learn $Q$ value that have close connection to shortest distance between two state and could be used to determine lengths of each edge in the planning graph.\\

\noindent\emph{Planning}: The idea of combing planning in RL has been around for a long time \cite{NIPS1992_714,SUTTON1999181}. A recent work \cite{vin} use CNN as planner which convey useful global information to reactive policy. Another work \cite{deeploco} use hierarchical RL, where high level controller set goals and low level controller produce locomotion. Both controllers are trained in an actor-critic process. In this paper , waypoints are filled in state place in advance \cite{sorb} rather than generated dynamically during the task. Since agent can estimate the distance between two waypoints, it can search a path without additional learning.\\

\noindent\emph{Pattern Search}: The two planning parameters is optimized according to the testing result of tasks. The optimization has no gradient. After changing parameters, the planning graph should also be changed which could take a long time. Therefore, pattern search is used to reduce optimization time. In the original pattern search method \cite{patternsearch}, searching interval is gradually reduced to precisely find the optimal value. In this paper, the searching interval is increased to quickly find the range of optimal value. Adjustments are made to ensure an appropriate termination of search.

\section{Algorithm}

The online adapting algorithm has two part: optimizing planning parameters and pattern search. The optimization strategy is derived from analysis of relationship between planning and reaction. Pattern search accelerate, give a soft convergence circumstance to, and set a termination condition on the optimization.

\subsection{Optimizing Planning Parameters}

The two planning parameters is changed according to three different testing results of tasks: agent reaches the goal successfully; agent finds a path to the goal but cannot reach it; agent cannot find a path to the goal.\\

\noindent\emph{Success}: A shortest path is got by visiting to waypoints in the planning graph using Dijkstra's Algorithm, which takes more time as the number of waypoints grows. If agent could reach the goal, we could try to set less waypoints to get a quicker reaction.\\

\noindent\emph{Cannot Reach}: This means there is a pair of adjacent waypoints in the path that agent cannot move from one to another by reaction. The shortest distance of two states is estimated by Q network trained through DDPG. This distance estimates is efficient but not accurate enough (See Appendix C for demonstration). In SoRB, three Q networks are trained together and distributional Q Values \cite{distributional-rl} are used to ensure robust distance estimates. If the problem still exist given these, the reaction capability of agent must be overestimated. Therefore, the maximum edge length of planning graph should decrease so that easier subgoals are set to the agent.\\

\noindent\emph{No Path}: This means the start and goal state are not connected in the planning graph. It is caused by sparsity of waypoints or edges. The solution is to add both two parameters which could bring more waypoints and edges into the graph.\\

\noindent Combing the above three situations, we can get an optimization algorithm, shown in Algorithm \ref{alg:oppo}.

\begin{algorithm}
	\caption{Optimizing Planning Parameters Once\\
		Inputs are a search policy $sp$, a rollout function $rf$ which tests the agent and returns frequencies of occurrence of each situation, and a threshold number $cth(=0.05)$ which set a condition on changing parameters.}
	\label{alg:oppo}
	\begin{algorithmic}
		\Function{UPDATE}{$sp, rf, cth$}
		\State $no\_problem \gets True$
		\State $result \gets rf(sp)$
		\If{$result.rate\_of\_cannot\_reach > cth$}
		\State the maximum edge length decreases
		\State $no\_problem \gets False$
		\ElsIf{$result.rate\_of\_no\_path > cth$}
		\State both two parameters increase
		\State $no\_problem \gets False$
		\EndIf
		\If{$no\_problem$}
		\State the number of waypoints decreases
		\EndIf
		\EndFunction
	\end{algorithmic}
\end{algorithm}

\subsection{Pattern Search}

The aim of using pattern search is to quickly determine the range of optimal value, and then narrow this range. The planning parameters may fluctuate to some extent but are close to the optimal value. Each parameter is optimized independently and an extra \emph{group of parameters} are used to record its optimization status. The process of pattern search is described below, taking the number of waypoints (denoted by $w$) as an example.

Initially, $w$ is set to a small enough value. The increment of $w$ (denoted by $i$) is larger than the decrement (denoted by $d$). The reason for such setting is that smaller $w$ is more likely to cause an failure which is dangerous while larger $w$ increase task time which is relatively tolerable. A larger $i$ could avoid $w$ from converging into an dangerous area.

At the beginning, $w$ increases continuously and $i$ increases exponentially which makes $w$ far exceed the optimal value. Then $w$ is set to its last searching value, $i$ is set to its initial value and $w$ will continue increasing. After several such repetition, the optimal value is determined within a small range. Now we can fix $i$ and try to reduce $w$. When \emph{'No Path'} happens, meaning that $w$ may enter the dangerous area, it should increase. Although agent may perform well in the following tasks, risk still exists. Therefore $d$ is reduced simultaneously to restrict attempt at reducing $w$. As $d$ decrease, $w$ gradually move away from dangerous area and fluctuate in a small area. The search is terminated when $d$ is small enough. A clear process is shown in Algorithm \ref{alg:ps}. To further accelerate the optimization, another search process is provided (See Appendix E for detail).

\begin{algorithm}
	\caption{Pattern Search\\
	Given number of exponential search times: $n(=3)$,\\
	increment of $w$: $i(=3)$, decrement of $w$: $d(=1)$,\\
	growth factor of $i$: $\rho(=2)$, reduction factor of $d$: $\gamma(=0.9)$,\\
	and termination threshold: $tth(=0.1)$,\\
	where $w$ (initialized to 1) denotes the number of waypoints.}
	\label{alg:ps}
	\begin{algorithmic}
		\State $k \gets 1$
		\State {\bfseries when} $w$ should increase :
		\State \indent $w \gets w+k\times i$
		\State \indent {\bfseries if} $n>0$ {\bfseries then}
		\State \indent \indent $k \gets k\times \rho$
		\State \indent {\bfseries else if} $n=0$ {\bfseries then}
		\State \indent \indent $d \gets d\times \gamma$
		\State \indent {\bfseries end if}
		\State {\bfseries when} $w$ should decrease :
		\State \indent {\bfseries if} $n>1$ {\bfseries then}
		\State \indent \indent $n\gets n-1; w\gets w-\frac{k}{\rho}\times i; k\gets 1$
		\State \indent {\bfseries else if} $n=1$ {\bfseries then}
		\State \indent \indent $n\gets n-1; w\gets w-d; k\gets 1$
		\State \indent {\bfseries else if} $n=0$ {\bfseries then}
		\State \indent \indent $w\gets w-d$
		\State \indent {\bfseries end if}
		\State {\bfseries when} $d<tth$ :
		\State \indent End search
	\end{algorithmic}
\end{algorithm}

\section{Experiment}

The experiments are taken in a 2D environment (See Appendix). First, a satisfactory parameter setting are got using the adapting algorithm. Then, one of the planning parameters is fixed to see the effect of another on the task time and success rate.

\subsection{Changing Process of Planning Parameters}

We first consider the situation when planning parameters converge. The meaning of parameters below can be found in Alorithm \ref{alg:oppo} and \ref{alg:ps} where one of their setting in the experiments is given.

When $w$ converges (concentrate on $w$ since maximum edge length is changed along with $w$), it fluctuates around a certain value. On average, it goes up $tth$ times every time it goes down $i$ times. When it goes up, \emph{'No Path'} happens in at least $cth$ of the tasks (assume the number is exactly $cth$). When it goes down, the frequency of \emph{'No Path'} is less than $cth$ (assume the number is exactly $cth/2$). Then we can calculate the success rate of task when $w$ is around its convergence value:
\begin{equation}
sr=1-\frac{i\times \frac{cth}{2}+tth\times cth}{i+tth} \label{equ:1}
\end{equation}
Notice that $i$ is much larger than both $tth$ and $cth$. Equation \ref{equ:1} can be simplified:
\begin{equation}
sr\approx 1-\frac{cth}{2} \label{equ:2}
\end{equation}
This means, using adapting algorithm, we would finally get a convergent $w$ value that make agent success in $(1-cth/2)$ of tasks. Such prediction is not accurate enough since it is derived under many assumptions, however, it is useful for understanding the training result.

\begin{figure}[H]
\centering
\subfigure[$i=10$]{
\includegraphics[width=.4\textwidth]{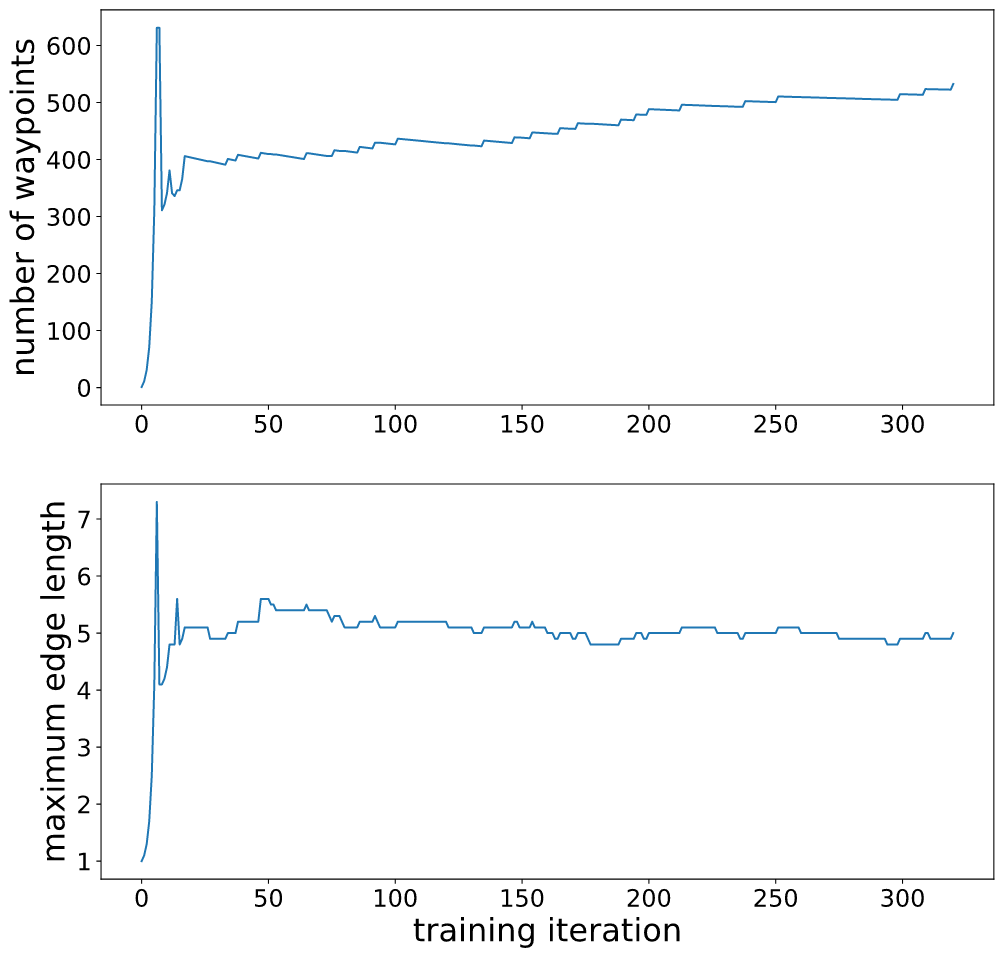}
\label{fig:i=10}
}
\subfigure[$i=3$]{
\includegraphics[width=.4\textwidth]{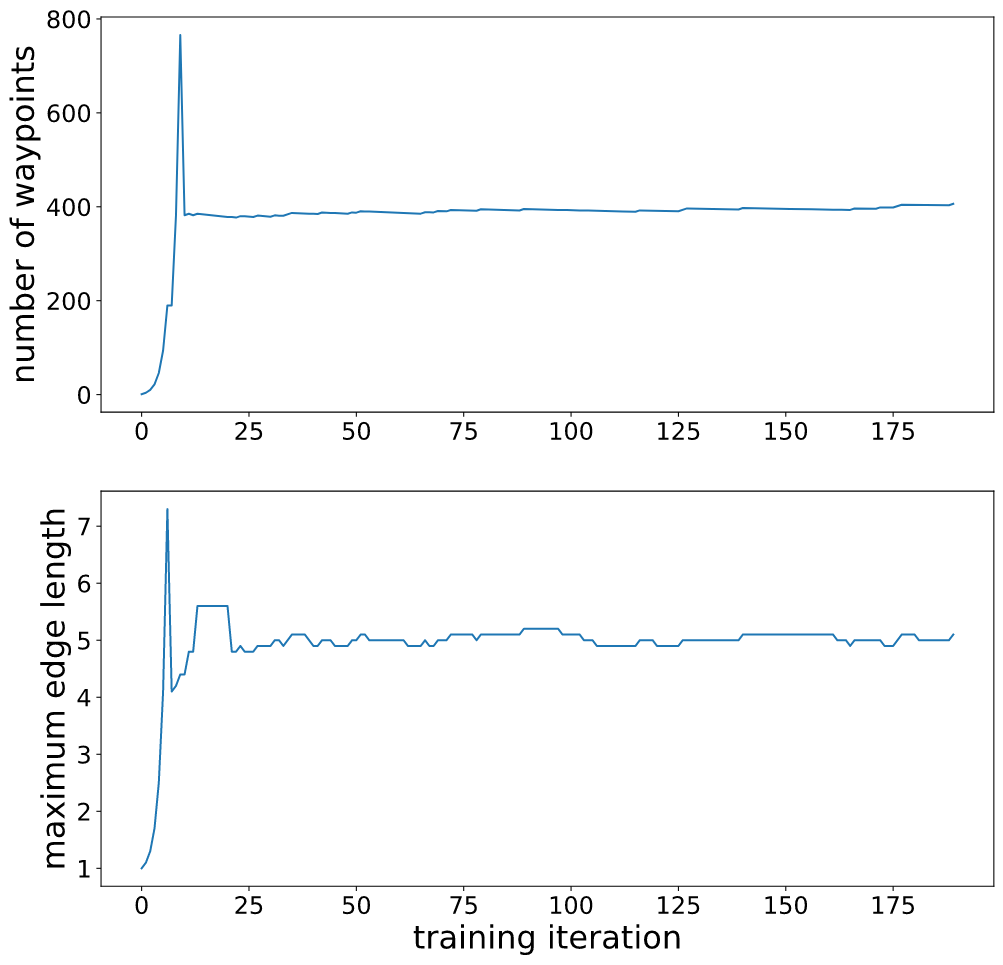}
\label{fig:i=3}
}
\caption{$(Left)$ The larger $w$, the higher success rate the agent will achieve. A larger $i$ would lead to higher success rate because it is more inclined to avoid failure. $(Right)$ For a smaller setting of $i$, $w$ gets stable around 400.}
\end{figure}

Two changing processes of planning parameters are shown. The setting of extra parameters in Figure \ref{fig:i=10} is same as those in Algorithm \ref{alg:oppo} and \ref{alg:ps} except that $i$ is set to 10. In Figure \ref{fig:i=3}, the setting is totally the same. Reaction capability also has a influence on the convergence value of $w$ (See Appendix D for detail).

The randomness comes from two parts: waypoints and tasks are randomly sampled. In each iteration, there are 40 different waypoints settings, in each of which 5 different tasks are given.

\subsection{Comparison of Different Planning Parameters Settings}

\begin{figure}[H]
\centering
\subfigure[different $w$]{
\includegraphics[width=.4\textwidth]{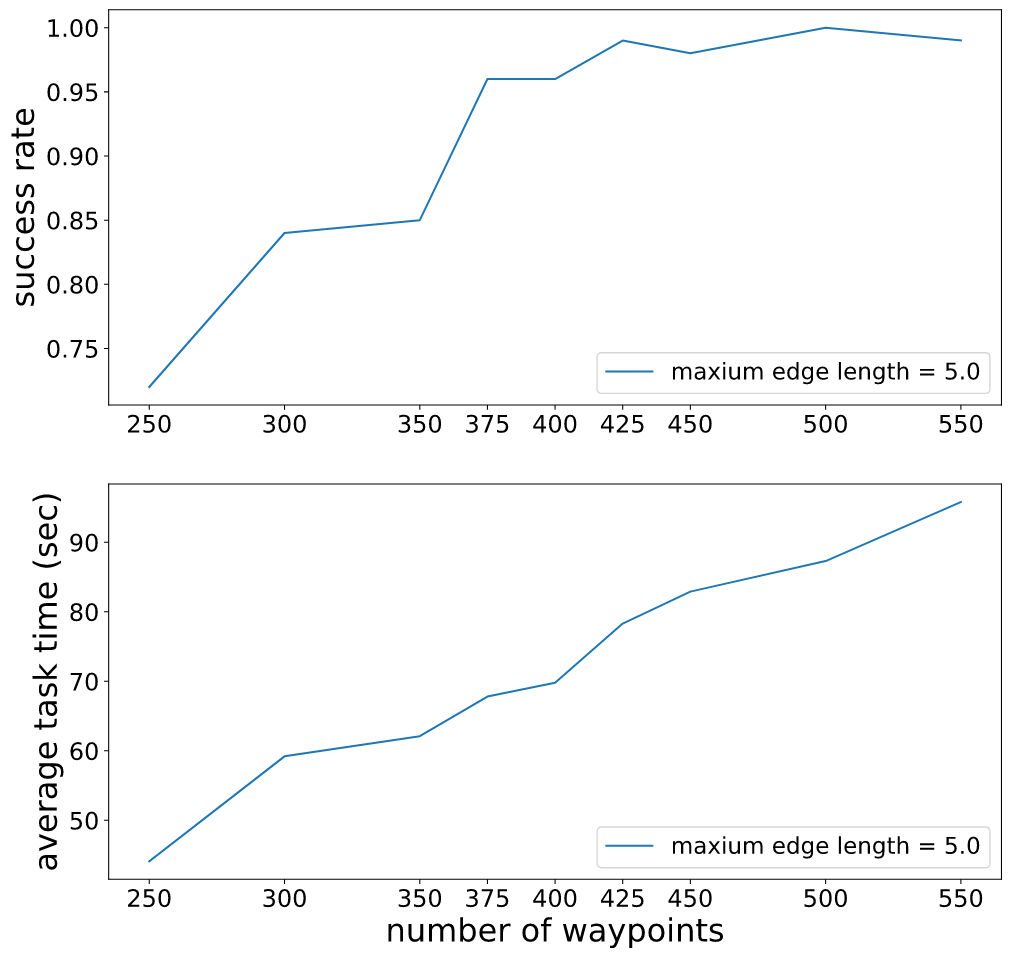}
\label{fig:cp_nw}
}
\subfigure[different $e$]{
\includegraphics[width=.4\textwidth]{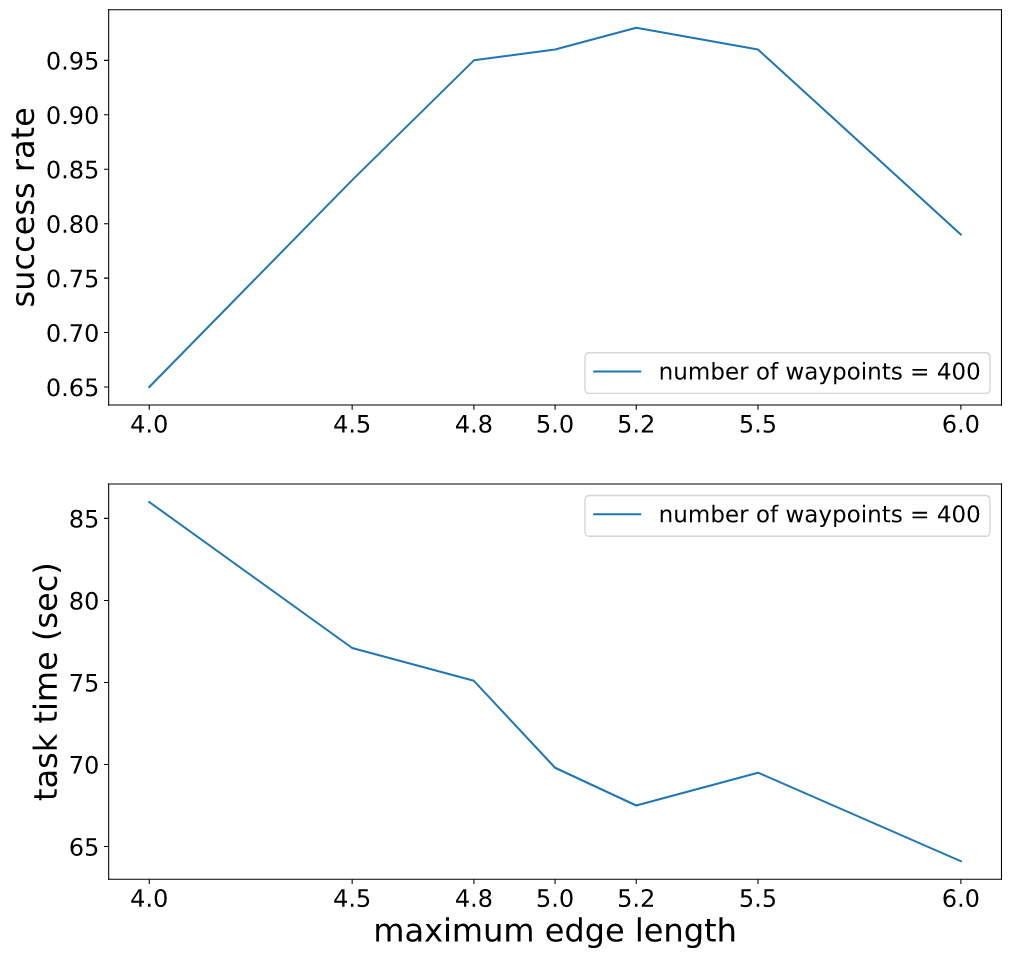}
\label{fig:cp_me}
}
\caption{$(Left)$ Success rate improves little when $w$ and itself exceed certain values. The $sr$ derived from Equation \ref{equ:1} is close to the certain value, so is the converged $w$. $(Right)$ Larger $e$ is more likely to cause occurrence of \emph{'Cannot Reach'}, but provide shorter paths, leading to less task time. Smaller $e$ is more likely to cause \emph{'No Path'}, and hide shortcuts, causing more task time and lower success rate.}
\end{figure}

In Figure \ref{fig:i=3} we get a satisfactory setting of planning parameters: $w=406.3$ and $e=5.1$, where $e$ denotes the maximum edge length of planning graph. Taking this setting as center, we now compare task time and success rate in different parameter settings. We first fix $e$ to 5 and change $w$. Then we fix $w$ to 400 and change $e$. The results are shown in Figure \ref{fig:cp_nw} and \ref{fig:cp_me}.

The distances of each pair of waypoints are cached before the task so that the time complexity of searching next waypoint is reduced from $o(w^2)$ to $o(w)$ \cite{sorb}. This makes task time grows almost linearly with $w$. Failed tasks are not counted when calculating average task time.\\

\noindent The experiments show that with appropriate parameter setting, the pattern search can quickly find a satisfactory setting of two planning parameters. This method can be extend to more sophisticated problems where there are more than three planning parameters to optimize without gradient, as long as the optimization strategy (similar to Algorithm \ref{alg:oppo}) is given.

\section{Discussion and Future Work}

Combining planning and reaction could help to handle complicated tasks which have high-dimensional data stream, long term decision process and sparse supervision signal. A good planning algorithm could make full use of limited reaction capability which is usually obtained by deep reinforcement learning. A specific planning algorithm has parameters that need changing to accommodate the reaction capability of agent. The optimization direction of these parameters could not derive from calculating gradients. Therefore, we need to figure out the relationship between planning and reaction to create optimization strategy. After determining the strategy, improved pattern search method can be used to greatly accelerate optimization.

In this paper, planning method creates a memory of state place where agent can get useful instructions during tasks. In the future, we could design a planner which create and remove waypoints repeatedly, to form a more efficient memory of the environment. We could also try to simultaneously improve planning and reaction, which might bring us powerful agents with excellent reactions (see Figure \ref{fig:turn}) in the whole environment. Furthermore, agent needs to explore the environment when there is no waypoint initially.

\bibliographystyle{plain}
\bibliography{ref}

\clearpage

\begin{appendices} 

\section{Introducing Planning}

\begin{figure}[h]
	\centering
	\includegraphics[width=8cm,height=4cm]{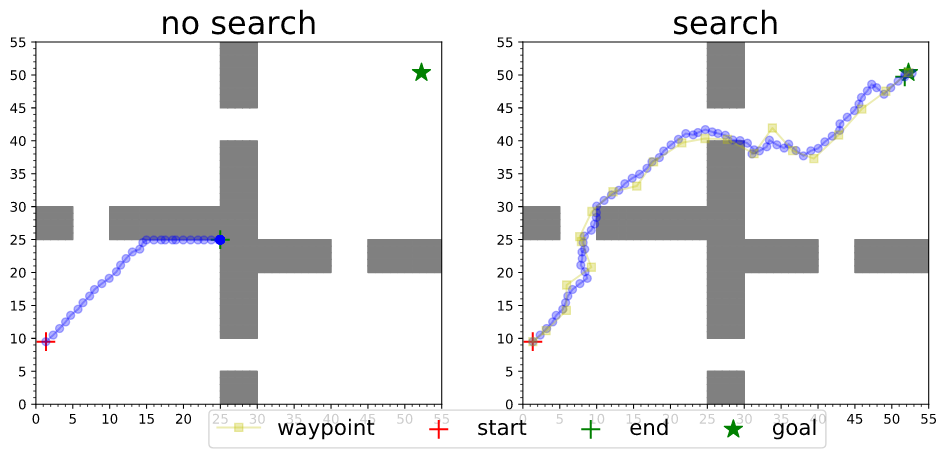}
	\caption{Setting waypoints in the environment, agent achieves the goal step by step.}
	\label{fig:s-and-ns}
\end{figure}

Figure \ref{fig:s-and-ns} shows the 2D environment where experiments are done. 

It is difficult to train a DDPG agent (left), because supervision signal only appears when agent is close to the goal and agent gets no reward in most of its experience. If we set waypoints (right) that agent can reach one by one, the remote goal is easier to achieve.

\begin{figure}[h]
	\centering
	\includegraphics[width=5cm,height=5cm]{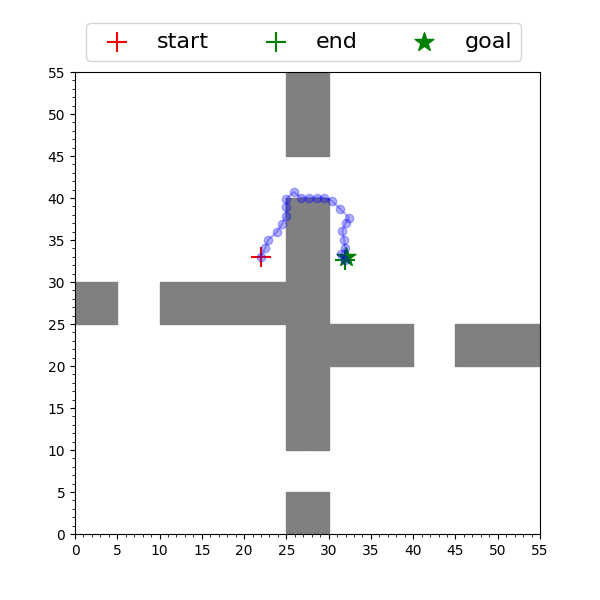}
	\caption{Using a special training recipe, agent can take excellent reaction locally.}
	\label{fig:turn}
\end{figure}

\section{Deliberate Training}

There are tricks for improving reaction capability substantially. However, such method could not generalize to the whole environment. In Figure \ref{fig:turn}, agent can reach the goal only by reacting, which means it remember the walls in some way (parameters in the reacting network).

To achieve this, agent is first trained in an simple environment where there are no walls. This could make it move in the right direction. The reacting network obtained is reused in this new environment \cite{transfer,unreal}. Then agent is repeatedly trained in a same task ($start$ and $goal$ are fixed) and the difficulty of task is gradually increasing (agent needs to bypass increasingly thicker walls). After such complicated training process, agent get excellent reaction in this small area.

To extend such ability to the whole environment, we need much more training data to let the reacting network get a memory of all the walls. The scale of such network is unknowable. Besides, training process should be set carefully to avoid catastrophic forgetting \cite{dqn}.

To get more efficient training data, imitation learning could be used \cite{Rahmatizadeh2016LearningRM}. This requires manually provided data which is got from human experience.

\begin{figure}[H]
	\centering
	\subfigure[Problematic Q Value]{
		\includegraphics[width=6cm,height=5cm]{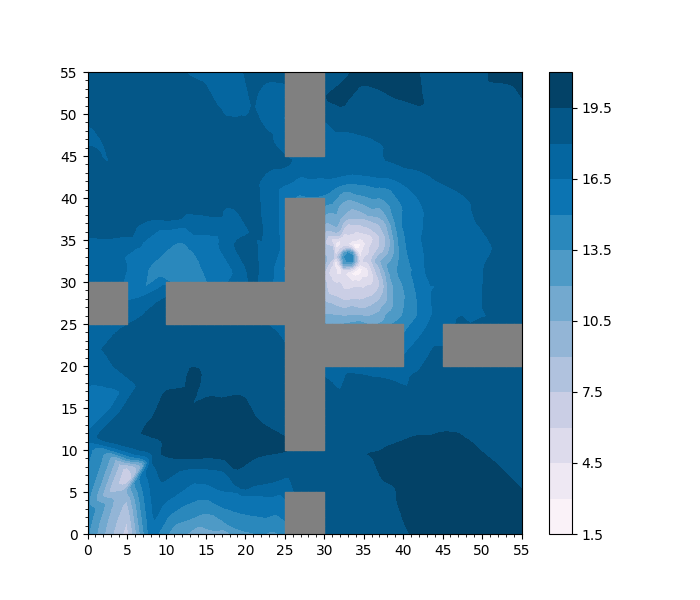}
		\label{fig:bad-dis}
	}
	\subfigure[Problematic Planning]{
		\includegraphics[width=4.5cm,height=4.5cm]{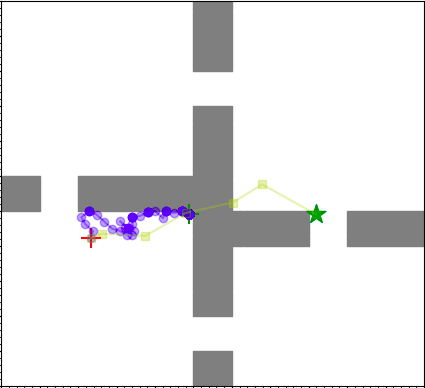}
		\label{fig:through}
	}
	\caption{$(Left)$ This is a problematic Q value. If there are two waypoints in (33,33) and (5,7), the planning graph will contain an edge connecting these two waypoints. $(Right)$ The distance of waypoints on both sides of the wall should have been large but is small. As a result, agent try to go through the wall.}
\end{figure}

\section{Problematic Distance Estimates}

The Q values learned by DDPG is accurate only in a small area that is around agent. This is enough, because the Q values is used to determine edge length of close waypoint pairs and reactive policy need not to care about remove state.

Figure \ref{fig:bad-dis} shows an image of all the Q values. (33,33) is the center.

In the goal-conditioned RL, an training episode end when the agent is close enough to the goal (in a center circle). Therefore, there is no transition where start state is in the center circle (i.e., start and goal state are very close), causing lack of training data and agnostic Q values nearing the center. Fortunately, this has not caused trouble in the experiments, because agent is not likely to choose a subgoal that is extremely close to it.

Another problem could cause much trouble: There exist some pairs of states whose shortest path is large but considered small (notice the white part in the corner of Figure \ref{fig:bad-dis}). Agent would add an edge to such pairs in the planning graph. But in fact, agent could not react from one to another. Figure \ref{fig:through} exemplifies a typical trouble.

\section{Comparison of Different Reaction Capability}

\begin{figure}[H]
\centering
\subfigure[trained with 2e5 steps]{
\includegraphics[width=0.4\textwidth]{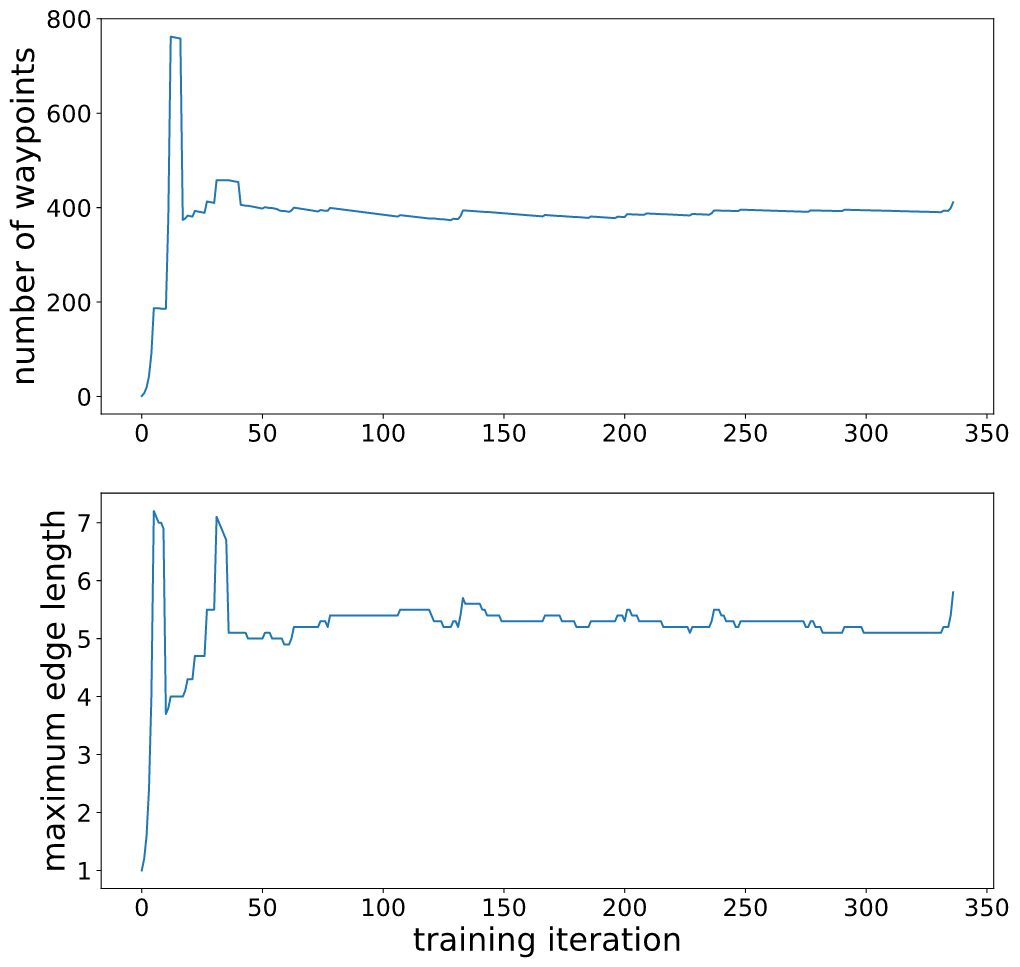}
\label{fig:20w-new}
}
\subfigure[trained with 4e4 steps]{
\includegraphics[width=0.4\textwidth]{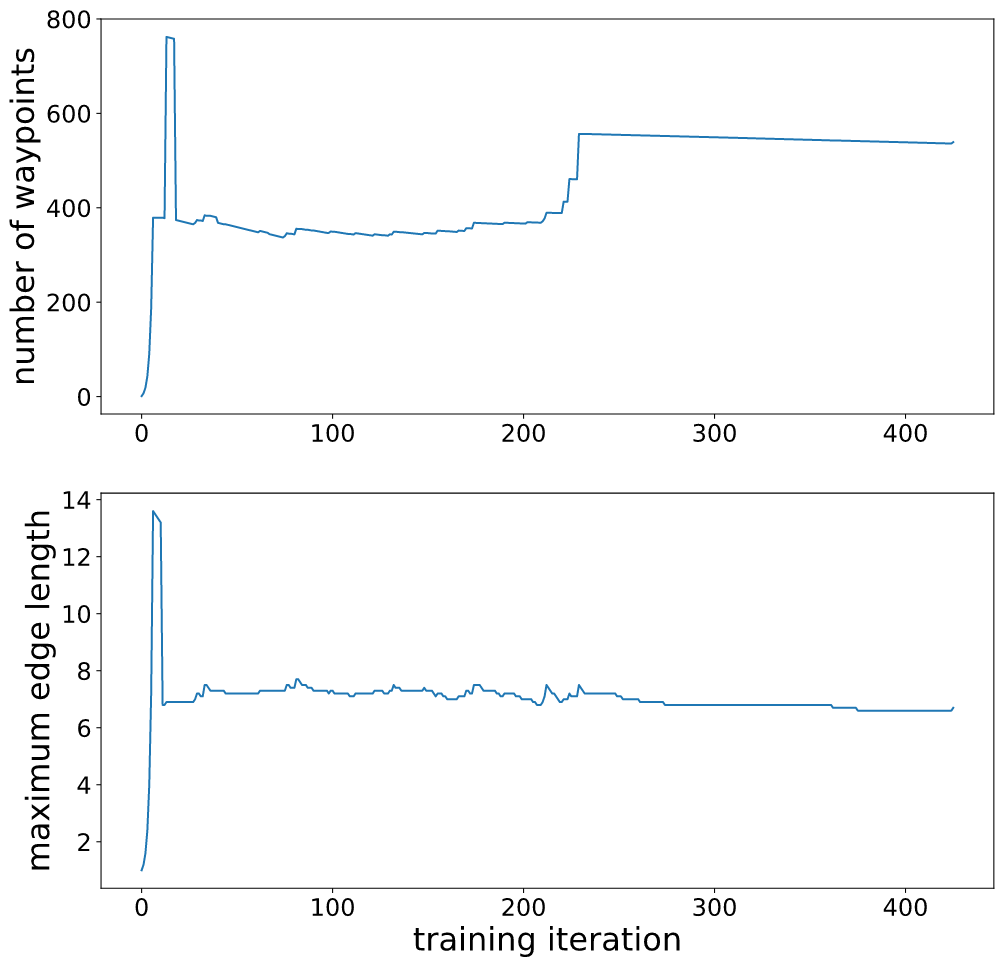}
\label{fig:4w-new}
}
\caption{$(Left)$ This is the same agent as one in Section 4 $(i=3)$. $w$ and $e$ fluctuate around 400 and 5.3. $(Right)$ This agent does not get enough reactive training. It needs more waypoints to ensure good performance.}
\end{figure}

Agents with different reaction capability have different requirement for planning parameters. The agent in Figure \ref{fig:20w-new} is the same as one in Section 4 which is trained with 200 thousand steps by DDPG, whereas the agent in Figure \ref{fig:4w-new} is only trained with 40 thousand steps.

These training curves has obvious difference to those in Section 4, because another pattern search algorithm (See Appendix E) is used to further accelerate the optimization of planning parameters. The randomness of experiments could hinder the optimization if we use Algorithm \ref{alg:ps}.

%

In Figure \ref{fig:4w-new}, $e$ fluctuates around 7 which conflicts with intuition. $e$ should have been smaller than the one in Figure \ref{fig:20w-new}, because an agent trained with fewer steps has worse reaction capability. The reason for a larger $e$ is that the distances estimated by this agent are generally larger. An agent performing badly often overestimates distances of two state. This phenomenon further reveals a characteristic of distance measurement: It is a heuristic create by agent within, and its fundamental purpose is to help agent make decisions rather than predictions.

\section{Another Pattern Search Method}

\begin{figure}[h]
	\centering
	\includegraphics[width=5cm,height=5cm]{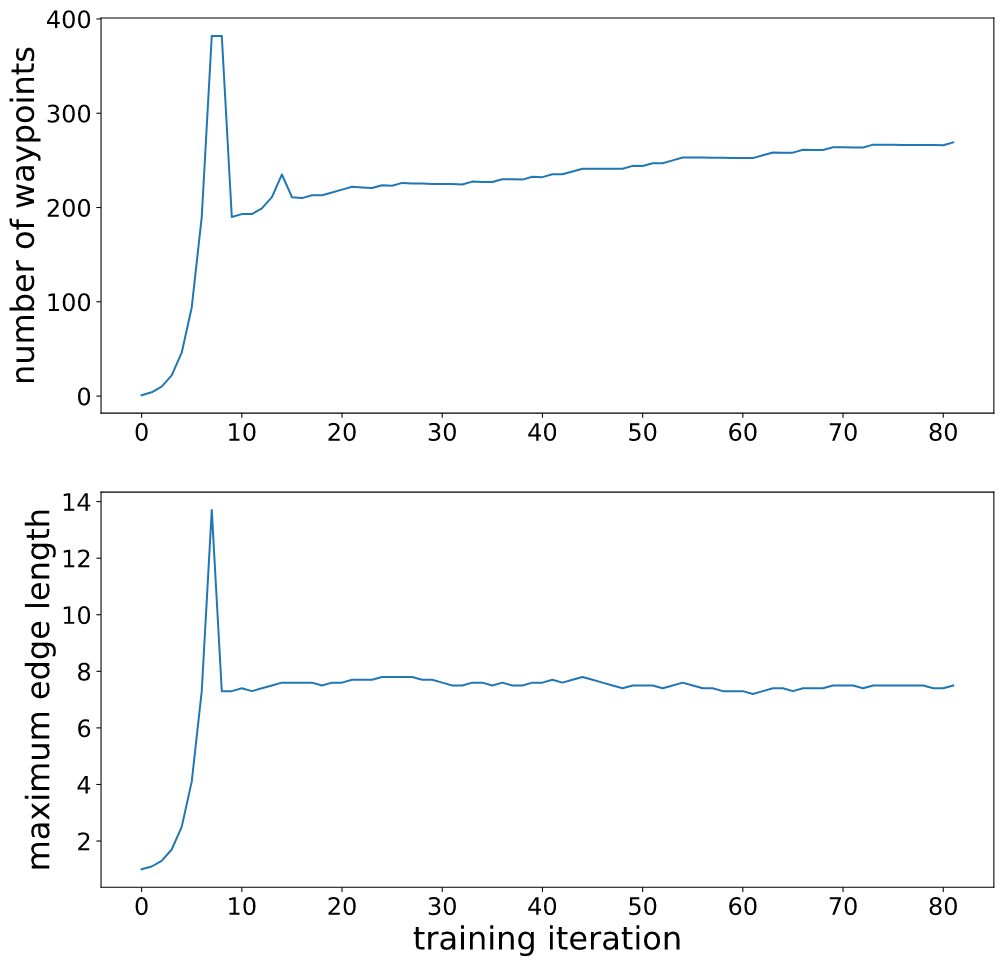}
	\caption{\emph{'Success'} happens at 8th and 14th iteration, causing early ending of exponential growth. Then $d$ continuously decrease as $w$ increase. Before $w$ rise to its convergence value, $d$ has fallen below the threshold, which cause an early ending. See Figure \ref{fig:4w-new} for comparison.}
	\label{fig:4w-old}
\end{figure}

In Algorithm \ref{alg:ps}, we set an end condition that could be fulfilled when $w$ increase enough times. If the algorithm ends normally, agent would get a high success rate. Even if $w$ is at a small value, there are still some probability for success, which would cause early ending of exponential growth and whole search. Figure \ref{fig:4w-old} gives an example.

For a small $w$, although we could not expect failures to occur one after another, there is a large frequency of them. To make $w$ increase quickly in such situation, we could extend the time span of pattern search, creating conditions for exponential growth that are easier to meet. In Algorithm \ref{alg:ps}, exponential growth happens when growth also happened on previous iteration. Now, the condition is not limited to last one but several iterations, and termination no longer happens. Algorithm \ref{alg:aps} shows the new pattern search process.

\begin{algorithm}
	\caption{Pattern Search\\
		Given number of exponential search times: $n(=3)$,\\
		increment of $w$: $i(=3)$, decrement of $w$: $d(=1)$,\\
		growth factor of $i$: $\rho(=2)$, reduction factor of $d$: $\gamma(=0.9)$,\\
		exponential search interval: $count(=4)$,\\
		and termination threshold: $tth(=0.1)$,\\
		where $w$ (initialized to 1) denotes the number of waypoints.}
	\label{alg:aps}
	\begin{algorithmic}
		\State $k \gets 1; c \gets 0$
		\State {\bfseries when} $w$ should increase :
			\State \indent {\bfseries if} $c>0$ {\bfseries then}
				\State \indent \indent $k \gets k\times \rho$
		\State \indent {\bfseries else if} $c=0$ {\bfseries then}
			\State \indent \indent $k \gets 1$
		\State \indent {\bfseries end if}
		\State \indent {\bfseries if} $n=0$ {\bfseries then}
			\State \indent \indent $d \gets d\times \gamma$
		\State \indent {\bfseries end if}
		\State \indent $w \gets w+k\times i$
		\State \indent $c \gets count$
		\State {\bfseries when} $w$ should decrease :
		\State \indent {\bfseries if} $n>0$ {\bfseries and} $c=1$ {\bfseries then}
		\State \indent \indent $n\gets n-1; w\gets w-k\times i$
		\State \indent {\bfseries else}
		\State \indent \indent $w\gets w-d$
		\State \indent {\bfseries end if}
		\State \indent {\bfseries if} $c>0$ {\bfseries then}
		\State \indent \indent $c \gets c-1$
		\State \indent {\bfseries end if}
		\State {\bfseries when} $d<tth$ :
		\State \indent End search
	\end{algorithmic}
\end{algorithm}

\section{Environment and Hyperparameters}

The 2D environment used in this paper is the same as one in SoRB \cite{sorb}, except that the noise of environment is smaller: the standard deviation of noise is 1.0 in SoRB and 0.3 here. Smaller environment noise makes task easier, and hence reduces required waypoints, making the training process shorter. Settings for RL training are list in Table \ref{table:p}.

\begin{table}
	\centering
	\begin{tabular}{c|c}
		{\bfseries Parameter} & {\bfseries Value}\\
		\hline
		learning rate & 1e-4\\
		training iterations & 2e5 and 4e4\\
		batch size & 64\\
		training steps per environment step & 1:1\\
		random steps at start of training & 1000\\
		replay buffer size & half of training iterations\\
		discount & 1\\
		OU-stddev, OU-damping & 1.0, 1.0\\
		target network update frequency & every 5 steps\\
		target network update rate & 0.05\\
	\end{tabular}
	\caption{Hyperparameters}
	\label{table:p}
\end{table}

\end{appendices} 

\end{document}